\documentclass[runningheads]{llncs}
\usepackage[T1]{fontenc}
\usepackage{graphicx}
\usepackage{amsmath}
\usepackage{amssymb}
\usepackage{booktabs}
\usepackage{xcolor}
\usepackage{hyperref}

\begin{document}

\title{Inferring Missing Trajectory Data with Temporal Convolutional Networks}

\author{Ilinca Tiriblecea\inst{1} \and
Gabriel Turinici\orcidID{0000-0003-2713-006X} 
\inst{2}
}%
\authorrunning{I.Tiriblecea \and G. Turinici}
\institute{Independent Researcher, Oxford, UK \and CEREMADE,  Universit\'e Paris Dauphine - PSL, CNRS, Paris, FRANCE \\
		\email{gabriel.turinici@dauphine.fr}, 
		\url{https://turinici.com} \\ \ \ \\ \ \ May 2026
        }

\maketitle
\begin{abstract}
Trajectory data collected in real-world settings is frequently incomplete due to
sensor failure, communication loss, or occlusion. We address the task of
\emph{trajectory inpainting}: reconstructing contiguous missing segments from
observed context. We propose a Temporal Convolutional Network (TCN) with
symmetric dilation that relaxes the standard causality constraint, allowing each
time step to draw on both past and future observations, a property that is
essential for inpainting, but absent from forecasting-oriented architectures.
The model is trained with a composite loss that combines weighted mean squared
error, boundary--continuity penalties, and a smoothness regularizer. Trained on a synthetic dataset of $1,000$ (train), $200$ (validation), and $300$ (test) two-dimensional trajectories with randomly
placed 20\% masked segments, the model achieves good R$^{2}$, MSE and MAE metrics.
\keywords{Trajectory inpainting \and Temporal Convolutional Networks \and
Sequence modeling \and Missing data reconstruction}
\end{abstract}

\section{Introduction}
\label{sec:intro}

Trajectory data underpins a broad class of real-world systems, from GPS-based
navigation to motion capture and autonomous robotics. In practice, however,
such data is rarely complete: drop-outs caused by signal obstruction, sensor
failure, or packet loss introduce contiguous gaps that must be filled before
any downstream processing can take place. The task of reconstructing plausible
trajectories from incomplete observations -- commonly referred to as
\emph{trajectory inpainting} -- requires a model that simultaneously respects
local continuity at gap boundaries and the broader temporal dynamics of the
full sequence.

Classical approaches such as linear interpolation are computationally
light\-weight but ignore the underlying motion model, producing unrealistic
straight-line segments that violate the natural curvature of the trajectory.
Recurrent models such as LSTMs~\cite{lstm1997} capture sequential dependencies
but suffer from sequential computation bottlenecks and vanishing gradients.
Transformer-based models provide strong long-range context but impose
substantial computational cost and typically require large training
corpora~\cite{vaswani2017attention}.

Temporal Convolutional Networks (TCNs)~\cite{bai2018empirical} offer an
attractive middle ground: sequences are processed entirely in parallel via
stacked dilated convolutions, training is stable, and the effective receptive
field grows exponentially with network depth. Crucially, because our objective
is inpainting rather than forecasting, we relax the standard causal-padding
constraint and adopt \emph{symmetric} padding, enabling every position to
attend to both past and future context. This design choice distinguishes our
use of TCNs from their conventional application to sequence prediction.

We evaluate the proposed approach on synthetic two-dimensional trajectories
composed of superimposed sinusoidal components with random phase offsets and
additive Gaussian noise, 
reporting performance on both a pilot and a larger-scale experimental setting.

\subsection{Literature review}

Pedestrian trajectory modeling has been extensively studied in the context of
autonomous driving, crowd analysis, and human–robot interaction, with the
primary focus historically placed on \emph{trajectory forecasting} rather than
reconstruction. Early approaches relied on expert-driven and physics-based
models, such as social-force formulations and heuristic crowd dynamics, which
offer interpretability but struggle with complex, nonlinear motion patterns
and long temporal dependencies~\cite{henderson1973crowd}.

The emergence of data-driven approaches shifted attention toward learning
trajectory dynamics directly from data. Recurrent neural networks, notably
LSTMs, became a dominant paradigm due to their ability to capture temporal
dependencies, with Social-LSTM~\cite{alahi2016social} introducing explicit
social interaction modeling via pooling mechanisms. Subsequent generative
extensions, such as Social-GAN~\cite{gupta2018social}, enabled multimodal
forecasting by sampling diverse plausible futures, significantly improving
performance on standard benchmarks such as ETH/UCY~\cite{pellegrini2009youll,lerner2007crowds}. However, these models are
primarily causal and forward-predictive, and are not explicitly designed to
handle missing observations within a sequence.

More recent work has explored convolutional and graph-based architectures to
overcome the limitations of recurrent models. Temporal Convolutional Networks
(TCNs) provide parallel computation and stable training while maintaining large
receptive fields through dilation~\cite{bai2018empirical,lea2016temporal}.
Graph-based models, including Social-STGCNN and related spatio-temporal graph
networks, explicitly encode pedestrian–pedestrian interactions and achieve
strong results in dense scenes~\cite{mohamed2020social,shi2021sgcn}. Attention
mechanisms and transformer-based models further improve long-range dependency
modeling, albeit at increased computational cost \cite{vaswani2017attention}.

Alongside architectural advances, benchmark datasets and evaluation protocols
have played a central role. ETH/UCY~\cite{pellegrini2009youll,lerner2007crowds} and the Stanford
Drone Dataset remain the most widely used references, typically evaluated using
displacement-based metrics such as ADE and FDE~\cite{gupta2018social,9710708}. However, as noted in recent
surveys~\cite{rudenko2020survey,taha_pedestrian_2026}, these datasets were not
originally designed for studying missing data. TrajImpute represents the
closest recent effort toward formalizing trajectory imputation by introducing
systematic masking protocols and imputation-focused evaluation~\cite{trajimpute}.

Overall, while trajectory prediction has seen rapid progress—particularly with
generative and diffusion-based models~\cite{sun2025gdts,bae2024singulartrajectory}—the
specific problem of trajectory inpainting remains comparatively underexplored.
Existing methods are often adapted post hoc to missing-data scenarios, and
there is a lack of standardized benchmarks isolating reconstruction from
forecasting. This explains our use of controlled synthetic datasets to study
trajectory inpainting in isolation, enabling precise analysis of architectural
and loss-function design choices.

\subsection{Summary and Positioning}

Trajectory inpainting remains comparatively underexplored relative to
forecasting and tracking, with fragmented benchmarks and largely adapted
objective formulations. This work makes several contributions: first, we introduce a controlled synthetic benchmark; second, 
we adapt the TCN architecture to trajectory inpainting by replacing causal convolutions with symmetric dilated convolutions that exploit both past and future context;
third, we design a loss function tailored to handle contiguous gaps.

\section{Methodology}
\label{sec:method}

\subsection{Synthetic Data Generation}

\paragraph{Motivation:}
There is no single standardized trajectory inpainting benchmark; TrajImpute is
the closest recent effort, and our synthetic data provides a complementary,
fully controlled testbed~\cite{trajimpute}. Existing evaluations therefore rely
on a combination of synthetic trajectories, repurposed pedestrian datasets, and
tracking benchmarks.

Synthetic datasets are widely used to study interpolation and imputation under
controlled conditions, as parametric and sinusoidal motion models generate
smooth but nonlinear trajectories that cannot be recovered by linear
interpolation~\cite{wan2023generative,splineinterp}. Our synthetic data follows this
established practice while enabling precise control over curvature, noise, and
gap structure.

Real-world datasets such as ETH/UCY~\cite{pellegrini2009youll,lerner2007crowds} are frequently adapted for inpainting by
artificially masking contiguous segments~\cite{pellegrini2009youll}. TrajImpute formalizes
this approach through standardized masking protocols and evaluation metrics
focused explicitly on imputation performance~\cite{trajimpute}. In contrast,
tracking benchmarks such as MOT16-20 contain natural gaps due to
occlusions~\cite{leal2015motchallenge}, but reconstruction quality is confounded with
detection noise and data association.

In this context, synthetic trajectories serve not as a replacement for real-world data, but as a principled experimental baseline that isolates the inpainting problem from confounding factors such as detection noise and identity ambiguity. This controlled setting enables precise analysis of model behavior and loss design, which is difficult to achieve on fully unconstrained benchmarks.

\paragraph{Technical details:}
Trajectories are generated as superpositions of sinusoidal components with
random phase offsets and additive Gaussian noise. For a normalized time axis
$t \in [0,1]$ sampled at $T = 200$ evenly spaced points, the $x$ and $y$
coordinates are defined as:
\begin{align}
  x(t) &= \sin(2\pi t + \phi_x) + 0.3\sin(6\pi t + \phi_y)
          + \epsilon_x, \label{eq:x}\\
  y(t) &= \cos(2\pi t + \phi_y) + 0.3\sin(4\pi t + 0.5 + \phi_x)
          + \epsilon_y, \label{eq:y}
\end{align}
where $\phi_x, \phi_y \sim \mathcal{U}(0, 2\pi)$ are independent random phase
shifts and $\epsilon_x, \epsilon_y \sim \mathcal{N}(0, \sigma^2)$ with
$\sigma = 0.02$. Phase randomization ensures that no two trajectories are
identical, while the multi-frequency structure produces non-trivial curvature
that is challenging to interpolate linearly.

A single contiguous masked segment covering $20\%$ of the sequence (i.e.\
40 time steps) is placed at a uniformly sampled starting index for each
trajectory. 
We generated 1,500 trajectory-mask pairs, of which $1{,}000$ were used for training, $200$ for validation and $300$ for testing.
\subsection{Input Representation}

Each time step $t_i$ is encoded as a four-dimensional vector:
\begin{equation}
  \mathbf{u}_i =
  \bigl[\,x_i \cdot (1 - m_i),\;\; y_i \cdot (1 - m_i),\;\; m_i,\;\;
  \tilde{t}_i\,\bigr]^\top,
\end{equation}
where $m_i \in \{0, 1\}$ is the binary mask (1 if the position is missing, 0
if observed), and $\tilde{t}_i = i / (T-1)$ is the normalized time index.
Zeroing the coordinates inside the gap prevents the model from observing ground
truth during training, while the explicit mask channel allows it to distinguish
missing from observed positions without relying on the magnitude of the input
signal. The time channel provides positional context that is otherwise absent
in a purely convolutional architecture.

\subsection{Model Architecture}

The network consists of five stacked TCN blocks followed by a $1{\times}1$
convolutional output head that projects to two output dimensions (predicted
$\hat{x}$ and $\hat{y}$). Each block applies the following operations in
sequence:

\begin{enumerate}
  \item A dilated 1D convolution with kernel size $k = 5$ and dilation
        $d_\ell = 2^{\ell-1}$ for block $\ell \in \{1,\ldots,5\}$, i.e.\
        dilations of 1, 2, 4, 8, and 16;
  \item Layer normalization across the channel dimension;
  \item ReLU activation;
  \item A residual connection that adds the block input to its output, using a
        $1{\times}1$ projection when input and output channel counts differ.
\end{enumerate}

All hidden layers use 64 channels. Symmetric (zero) padding is applied rather
than the causal padding used in forecasting TCNs, so that each position
attends to an equal window of past and future context.
The receptive field spans 125 time steps, allowing substantial contextual information to be incorporated from both sides of the gap.
Table~\ref{tab:arch} summarizes the NN architecture.

\begin{table}[t]
\centering
\begin{tabular}{|c|c|c|c|c|}
\toprule
\textbf{Layer} & \textbf{Type} & \textbf{Channels} & \textbf{Dilation}
               & \textbf{Purpose} \\
\midrule
    &                               &   &  & Position \\
Input   & ---                              & 4  & --- & + mask\\
    &                               &   &  & +time \\
\midrule
Block 1 & Conv1D + LN        & 64 & 1   & Local \\
        & + ReLU + Residual &    &     & features \\
\midrule
Block 2 & Conv1D + LN        & 64 & 2   & Medium \\
        & + ReLU + Residual &    &     & context \\
\midrule
Block 3 & Conv1D + LN        & 64 & 4   & Longer \\
        & + ReLU + Residual &    &     & dependencies \\
\midrule
Block 4 & Conv1D + LN        & 64 & 8   & Broader \\
        & + ReLU + Residual &    &     & patterns \\
\midrule
Block 5 & Conv1D + LN        & 64 & 16  & Global \\
        & + ReLU + Residual &    &     & structure \\
\midrule
Head    & Conv1D ($1{\times}1$) & 2  & --  & Predicts \\
        &                        &    &     & $(\hat{x}, \hat{y})$ \\
\bottomrule
\end{tabular}
\vspace{5pt}
\caption{Model architecture overview. }
\label{tab:arch}
\end{table}

\subsection{Loss Function}

Most prior learning-based approaches optimize a pointwise reconstruction loss,
typically mean squared error, over missing trajectory points. While effective
for minimizing average error, such objectives alone do not enforce continuity
or smoothness and often lead to visually implausible reconstructions,
especially at gap boundaries.

To address this, a number of previous works introduce auxiliary regularization terms,
such as velocity or acceleration penalties, or constraints on endpoint
consistency. These terms encourage smoother motion but are often applied
globally, without explicitly distinguishing between observed and missing
regions.

The loss function used in this work is designed to reflect the specific
requirements of trajectory inpainting:
\begin{enumerate}
    \item 
First, the reconstruction loss (here $\mathcal{L}_{MSE}$) is
weighted to emphasize accuracy within masked regions while still weakly
anchoring predictions in observed segments. This prevents trivial solutions in
which the model alters observed points to reduce global error.
\item
Second, an
explicit continuity loss (here $\mathcal{L}_{cont}$)  penalizes mismatches at the entry and exit of the
masked segment, directly targeting the most perceptually salient failure mode
of inpainting models. 
\item
Finally, a first-order smoothness penalty (here $\mathcal{L}_{smooth}$)  encourages
globally coherent motion without imposing a rigid parametric model.
\end{enumerate}

By decomposing the objective into region-aware reconstruction, boundary
continuity, and global smoothness terms, our formulation differs from prior
approaches that rely primarily on uniform MSE or generic temporal regularizers,
and is specifically aligned with the structural constraints of the inpainting
task. Training minimizes the composite objective:
\begin{equation}
  \mathcal{L} = \mathcal{L}_{\mathrm{MSE}}
              + \lambda_{\mathrm{cont}}\,\mathcal{L}_{\mathrm{cont}}
              + \lambda_{\mathrm{smooth}}\,\mathcal{L}_{\mathrm{smooth}},
  \label{eq:loss}
\end{equation}
with $\lambda_{\mathrm{cont}} = \lambda_{\mathrm{smooth}} = 0.5$.
We now describe the three terms of the overall loss function.
\paragraph{Weighted MSE.}
The reconstruction term penalizes errors in masked regions more heavily than
those in observed regions:

\begin{equation}
  \mathcal{L}_{\mathrm{MSE}} =
    \frac{1}{|\mathcal{M}|} \sum_{i \in \mathcal{M}}
      \|\hat{\mathbf{p}}_i - \mathbf{p}_i\|^2
    + \frac{\alpha}{|\bar{\mathcal{M}}|} \sum_{i \notin \mathcal{M}}
      \|\hat{\mathbf{p}}_i - \mathbf{p}_i\|^2,
\end{equation}
where $\mathcal{M}$ denotes the set of masked indices, $\mathbf{p}_i =
(x_i, y_i)^\top$ is the ground-truth position, $\hat{\mathbf{p}}_i$ is the
model prediction, and $\alpha = 0.1$.

\paragraph{Continuity loss.}
To prevent discontinuities at gap boundaries, we penalize the squared distance
between the prediction at the gap start point and the observed position
immediately before the gap, and between the gap end point and the observed
position immediately after:

\begin{equation}
  \mathcal{L}_{\mathrm{cont}} =
    \|\hat{\mathbf{p}}_{s} - \mathbf{p}_{s-1}\|^2
    + \|\hat{\mathbf{p}}_{e} - \mathbf{p}_{e+1}\|^2,
\end{equation}
where $s$ and $e$ denote the start and end indices of the masked segment,
respectively.

\paragraph{Smoothness loss.}
A first-order finite-difference penalty (mimicking the $H^1$ seminorm) discourages large frame-to-frame
changes in the predicted output:

\begin{equation}
  \mathcal{L}_{\mathrm{smooth}} =
    \frac{1}{T-1} \sum_{i=1}^{T-1}
      \|\hat{\mathbf{p}}_{i} - \hat{\mathbf{p}}_{i-1}\|^2.
\end{equation}

\subsection{Training Protocol}

The model is optimized with Adam~\cite{kingma2014adam} (learning rate
$3\times10^{-3}$) for 50 epochs. 
We use a batch size of $1$ because the dataset is small enough and larger batch sizes would lead to too few updates per epoch.
Experiments with larger batch sizes (up to 256) yielded similar final performance but required more epochs to converge.

\section{Results and Discussion}
\label{sec:results}

\subsection{Quantitative Evaluation}
The quantitative evaluation is based on three metrics described below. 
\paragraph{Evaluation metrics.}
Let $\{\mathbf{p}_i\}_{i \in \mathcal{M}}$ with
$\mathbf{p}_i = (x_i, y_i)^\top$ denote the ground-truth trajectory coordinates
over the masked index set $\mathcal{M}$, and
$\{\hat{\mathbf{p}}_i\}_{i \in \mathcal{M}}$ their reconstructions, with
$|\mathcal{M}| = T_m$. Let
$\bar{\mathbf{p}} = \frac{1}{T_m}\sum_{i \in \mathcal{M}} \mathbf{p}_i$ denote the
mean position over the masked segment.

The mean squared error (MSE) over masked regions,
the mean absolute error (MAE) and 
the coefficient of determination ($R^2$) are computed as
\begin{align}
\mathrm{MSE} &=
\frac{1}{T_m} \sum_{i \in \mathcal{M}}
\left\lVert \mathbf{p}_i - \hat{\mathbf{p}}_i \right\rVert_2^2,
\\
\mathrm{MAE} &=
\frac{1}{T_m} \sum_{i \in \mathcal{M}}
\left\lVert \mathbf{p}_i - \hat{\mathbf{p}}_i \right\rVert_1,
\\
R^2 &=
1 -
\frac{\sum_{i \in \mathcal{M}}
\left\lVert \mathbf{p}_i - \hat{\mathbf{p}}_i \right\rVert_2^2}
{\sum_{i \in \mathcal{M}}
\left\lVert \mathbf{p}_i - \bar{\mathbf{p}} \right\rVert_2^2}.
\end{align}

\paragraph{Interpretation.}
While MSE and MAE quantify absolute reconstruction error, $R^2$ measures how much
of the intrinsic variability of the masked trajectory segments is explained by
the model, providing a scale-independent assessment of structural fidelity. The
relatively large variance of $R^2$ across trials is attributable to the random
placement of masked segments, which induces substantial differences in
reconstruction difficulty depending on local curvature, frequency content, and
temporal context.

\paragraph{Baseline}
To compare the results with a simple baseline we also implemented a linear interpolation procedure.

\begin{figure}[ht]
\centering
\begin{minipage}{0.34\textwidth}
\centering
\begin{tabular}{c||c}
\toprule
\multicolumn{2}{c}{\textbf{Baseline: }} \\
\multicolumn{2}{c}{\textbf{Linear interpolation}} \\
\midrule
\midrule
\textbf{Metric} & \textbf{Mean $\pm$ Std} \\
\midrule
MSE & $0.047 \pm 0.096$ \\
MAE & $0.159 \pm 0.083$ \\
R$^2$ & $-0.437 \pm 5.639$ \\
\bottomrule
\end{tabular}
\\
\vspace{5pt}
\begin{tabular}{c||c}
\toprule
\multicolumn{2}{c}{\textbf{TCN model}} \\
\midrule
\midrule
\textbf{Metric} & \textbf{Mean $\pm$ Std} \\
\midrule
MSE & $0.004 \pm 0.012$ \\
MAE & $0.047 \pm 0.025$ \\
R$^2$ & $0.776 \pm 0.777$ \\
\bottomrule
\end{tabular}
\end{minipage}
\begin{minipage}{0.65\textwidth}
\centering
\includegraphics[width=\linewidth]{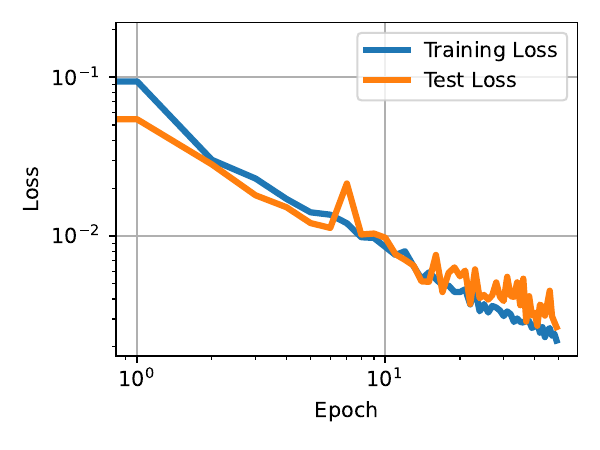}
\end{minipage}
\caption{
{\bf Left top:} Quantitative evaluation on masked regions (trained on $1{,}000$ synthetic
trajectories, 20\% mask fraction) for a linear interpolation baseline.
{\bf Left bottom:} Quantitative evaluation on masked regions (trained on $1{,}000$ synthetic
trajectories, 20\% mask fraction).
{\bf Right:} Loss evolution on the train and test datasets.        
}
\label{fig:nominal_run}
\end{figure}

\paragraph{Results}
Performance is evaluated exclusively on the masked (held-out) regions of each
trajectory. 
We report in Figure~\ref{fig:nominal_run}  MSE, MAE, and $R^2$, each
computed per trajectory and then averaged over the full $1,000$-sample dataset (together with the standard deviation over the trajectories).
The same figures are given for the baseline algorithm.

The $R^2$ figure indicates that the model captures most of the dominant
temporal structure of the synthetic trajectories. The associated standard
deviation reflects the heterogeneity of masked segments, with
performance degrading in cases where the gap overlaps regions of high local
curvature or rapid directional change. The model shows clear improvement with respect to the baseline in all  three metrics.

For comparison, an earlier experiment conducted on a smaller, less varied
dataset ($200$ trajectories drawn from only two underlying trajectory shapes)
yielded considerably stronger metrics, as shown in Table~\ref{table:results_pilot}.

\begin{table}[t]
\centering
\begin{tabular}{c||c||c}
\toprule
\textbf{Dataset} & \textbf{Metric} & \textbf{Mean $\pm$ Std} \\
\midrule
Smooth  & MSE & $0.0011 \pm 0.0005$ \\
        & MAE & $0.025  \pm 0.006$  \\
        & R$^2$ & $0.984 \pm 0.006$ \\
\bottomrule
\end{tabular}
\ \ 
\begin{tabular}{c||c||c}
\toprule
\textbf{Dataset} & \textbf{Metric} & \textbf{Mean $\pm$ Std} \\
\midrule
Wiggly  & MSE & $0.0025 \pm 0.0009$ \\
        & MAE & $0.038  \pm 0.007$  \\
        & R$^2$ & $0.965 \pm 0.011$ \\
\bottomrule
\end{tabular}
\vspace{5pt}
\caption{Pilot experiment results on the two-trajectory dataset (trained on $200$ samples).}
\label{table:results_pilot}
\end{table}

\begin{figure}[tb]
    \centering
    \includegraphics[width=0.99\linewidth]{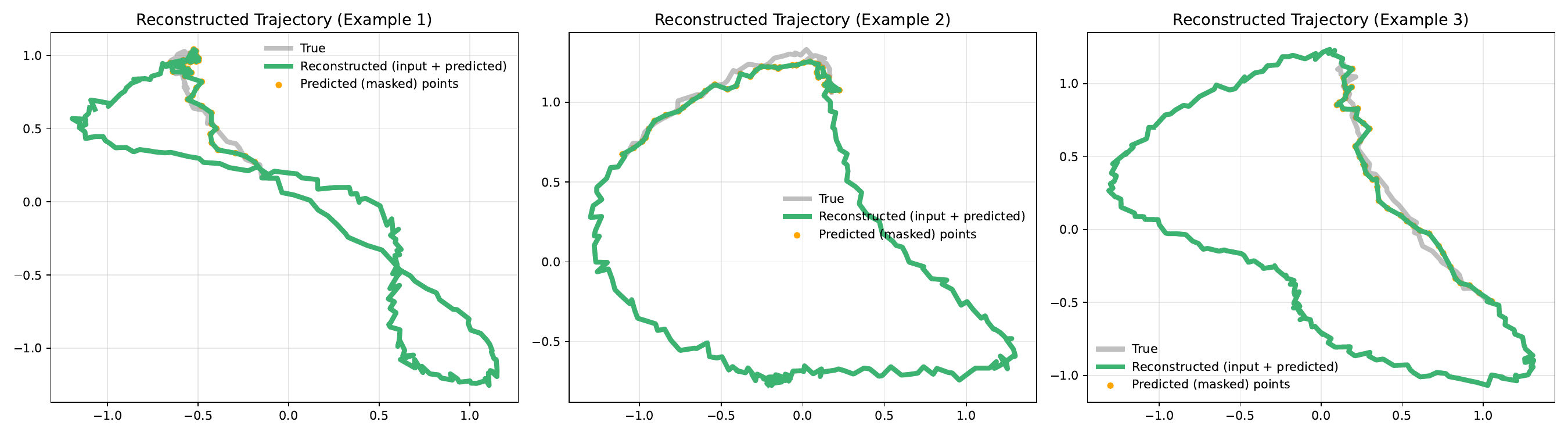}
    \caption{True versus predicted trajectories. The predicted trajectories are constructed using true values for the known part and predicted values for the masked part.}
    \label{fig:trajectories}
\end{figure}

The performance gap between the pilot and the full experiment is consistent
with the expectation that the 1{,}000-trajectory dataset introduces harder,
more diverse examples that expose the limits of the current architecture and
training configuration. We also examined the median of the per-trajectory scores as a
more outlier-robust summary; while it differs from the mean owing to a skewed
error distribution, it preserves the same ordering, with our model
outperforming the baseline on all three metrics.

\subsection{Qualitative Evaluation}

Qualitative inspection of reconstructed trajectories confirms that the model
produces smooth, boundary-consistent completions in the majority of cases.
The composite loss function visibly suppresses discontinuities at gap entry
and exit points relative to a pure MSE baseline, and the smoothness term
prevents high-frequency artifacts in the predicted segment. In cases where
quantitative performance is lower, visual inspection suggests that the model
tends to predict a smoothed, lower-amplitude version of the true trajectory
rather than an entirely erroneous path. Performance degrades when masked segments coincide with high-curvature regions,
highlighting the sensitivity of convolutional models to local frequency content.

\subsection{Ablation Study}

\begin{table}[t]
\centering
\begin{tabular}{c||c}
\toprule
\textbf{Metric} & \textbf{Mean $\pm$ Std} \\
\midrule
MSE & $0.005 \pm 0.018$ \\
MAE & $0.048 \pm 0.031$ \\
R$^2$ & $0.743 \pm 0.831$ \\
\bottomrule
\end{tabular}\ \ 
\begin{tabular}{c||c}
\toprule
\textbf{Metric} & \textbf{Mean $\pm$ Std} \\
\midrule
MSE & $0.005 \pm 0.013$ \\
MAE & $0.049 \pm 0.029$ \\
R$^2$ & $0.752 \pm 0.908$ \\
\bottomrule
\end{tabular}
\vspace{5pt}
\caption{Ablation study on loss function components (test set, 300 samples). 
{\bf Left:} no continuity loss. 
{\bf Right:}  no smoothness loss.
}
\label{table:ablation}
\end{table}

To assess the contribution of each loss component, we train two ablated
variants of the model: one without the continuity loss ($\lambda_{\mathrm{cont}}
= 0$) and one without the smoothness loss ($\lambda_{\mathrm{smooth}} = 0$),
keeping all other hyperparameters identical and using the same dataset. 
Results reported in Table~\ref{table:ablation} suggest a consistent, albeit modest, degradation across all metrics when either loss component is removed.
This conclusion depends, of course, on the specific run of the algorithm (because Adam is a stochastic optimization procedure).

\subsection{Limitations}

All experiments are conducted on synthetic data; generalization to real-world
trajectories -- which exhibit non-stationary dynamics, structured measurement
noise, and variable gap lengths -- remains an open question. The current
masking strategy places a single contiguous gap per trajectory; realistic
dropout patterns may involve multiple concurrent gaps. In addition, the fixed
receptive field of the TCN may require architectural scaling for substantially
longer input sequences. Finally, the difference between the mean and median scores
reflects a skewed error distribution in which a small number of hard
trajectories inflate the reported means; techniques that reduce this
variance across trajectories would be a valuable addition.

\section{Conclusion}
\label{sec:conclusion}

We have presented a TCN-based approach to two-dimensional trajectory
inpainting that combines a non-causal dilated convolutional architecture with
a composite loss function enforcing boundary continuity and output smoothness.
On a synthetic benchmark of 1{,}000 diverse trajectories the model achieves
a mean R$^2$ of $0.776$ on held-out masked regions, demonstrating that TCNs
are a viable, computationally efficient alternative to recurrent and
attention-based models for this task.

Future work includes evaluation on real trajectory traces, extension to multi-gap
masking, and a systematic ablation of the loss components to quantify the
individual contribution of the continuity and smoothness terms.

\bibliographystyle{splncs04}
\bibliography{references}
\end{document}